\documentclass{ieeeaccess}
\usepackage{cite}
\usepackage{amsmath,amssymb,amsfonts}
\usepackage{algorithmic}
\usepackage{algorithm}
\usepackage{graphicx}
\usepackage{textcomp}
\usepackage{caption}
\def\BibTeX{{\rm B\kern-.05em{\sc i\kern-.025em b}\kern-.08em
    T\kern-.1667em\lower.7ex\hbox{E}\kern-.125emX}}
\begin{document}
\history{Date of publication xxxx 00, 0000, date of current version xxxx 00, 0000.}
\doi{10.1109/ACCESS.2020.DOI}

\title{Auto-Ensemble: An Adaptive Learning Rate Scheduling based Deep Learning Model Ensembling}
\author{\uppercase{Jun Yang},
\uppercase{Fei Wang}}
\address{School of Electronic Information and Communications(EIC) from Huazhong University of Science and Technology (HUST), Luoyu Road 1037, Wuhan, China }

\markboth
{Author \headeretal: Auto-Ensemble: An Adaptive Learning Rate Scheduling based Deep Learning Model Ensembling}
{Author \headeretal: Auto-Ensemble: An Adaptive Learning Rate Scheduling based Deep Learning Model Ensembling}

\corresp{Corresponding author: Fei Wang (e-mail: wangfei@hust.edu.cn).}

\begin{abstract}
Ensembling deep learning models is a shortcut to promote its implementation in new scenarios, which can avoid tuning neural networks, losses and training algorithms from scratch. However, it is difficult to collect sufficient accurate and diverse models through once training. This paper proposes Auto-Ensemble (AE) to collect checkpoints of deep learning model and ensemble them automatically by adaptive learning rate scheduling algorithm. The advantage of this method is to make the model converge to various local optima by scheduling the learning rate in once training. When the number of local optimal solutions tends to be saturated, all the collected checkpoints are used for ensemble. Our method is universal, it can be applied to various scenarios. Experiment results on multiple datasets and neural networks demonstrate it is effective and competitive, especially on few-shot learning. Besides, we proposed a method to measure the distance among models. Then we can ensure the accuracy and diversity of collected models.
\end{abstract}


\titlepgskip=-15pt

\maketitle

\section{Introduction}
\label{sec:introduction}
\label{intro}
Optimizing structure of network and loss function is a NP-hard process \cite{b1}. To enhance the generalization capabilities of models, different network structures have been designed to apply to different scenarios. Hence, manually designed network structures are often highly targeted. According to different tasks, it often requires to redesign or optimize the network structure deeply to maintain generalization performance in the new scenarios, which spend a large amount of manpower and computing resources. Therefore, Neural Architecture Search (NAS) is proposed as a new method to construct powerful models. NAS searches network structure automatically and frees up expert time. However, NAS must consume large training budget to acquire the “best” network structure. Furthermore, NAS cannot guarantee the performance and generalization of model in the aspect of loss function and training algorithm. So, it only applies in the field of routine supervised learning with a large amount of labeled data \cite{b2,b3,b4,b5}. Hence, NAS is rare to involve in other machine learning fields like semi-supervised learning, few-shot learning, etc. Besides, fine-tune can also achieve better generalize, which requires significant expertise. The above methods are not universal enough, for the same problem, ensemble learning is widely used to solve the problem of accuracy and generalization in machine learning applications \cite{b6}. Traditional ensemble learning such as Random Forest and AdaBoost which can hardly extract features about a certain task. And the feature engineering is highly based on manual selection.To avoid huge training budget and complicated feature engineering, this paper attempts to provide a deep learning based simple and automatic ensemble method, which called Auto-Ensemble (AE), to improve performance and generalization. The key idea is by scheduling learning rate to automatically collect checkpoints of model and ensembling them in once training.

Adaptive ensembles like AdaNet, make use of NAS models and automatically search over a space of candidate ensembles\cite{b8}. Auto-Ensemble(AE) differs from it but is also an auto searching process. By scheduling the learning rate, AE searches in the loss surface to collect checkpoints of model for ensemble. Compared to AdaNet, it reduces computing resources, has considerable improvement and it’s easier to implement.

AE makes use of adaptive cyclic learning rate strategy to achieve Auto-Ensemble. Cyclic learning rate strategy takes advantage of SGD's ability to avoid or even escape false saddle points and local minima, by simply scheduling the learning rate, it can make the model converge to a better local optimal solution in a shorter iteration period \cite{b9}. However, cyclic learning rate requires manual setting of many hyperparameters, cannot guarantee the diversity of the collected checkpoints of model. Based on this, we used an adaptive learning rate strategy with less hyperparameters, which can automatically collect as many checkpoints of model as possible with high accuracy and diversity in once training. In order to ensure the diversity of the checkpoints collected each time, we proposed a method to measure the distance between checkpoints, which can ensure that the model converges to different local optimal solutions in the process of continuous training.

The main contributions of this paper are:

•	An easy-to-implement methodology for ensembling models automatically in once training. In addition to traditional supervised learning, our experiment on few-shot learning demonstrates that Auto-Ensemble can apply to other deep learning scenarios.

•	We proposed a method to measure the diversity among models, by which we can ensure the accuracy and diversity of models in the training process.

•	Our experiment demonstrates the efficiency of Auto-Ensemble. The accuracy of the classifier can be significantly improved and greatly exceed single model. These greatly reduces the workload of manual designed and optimized network models.

We organize the paper by first describe the significance and overview of Auto-Ensemble method. In the following section we briefly introduce the related work of our methodology. And explain each section of Auto-Ensemble method in detail in Section 3. In Section 4 we demonstrate our setup and the results in experimental procedures. In Section 5, we conclude the advantages and the future work in our research.
\begin{figure*}[t!]
  \centering
  \includegraphics[width=\textwidth]{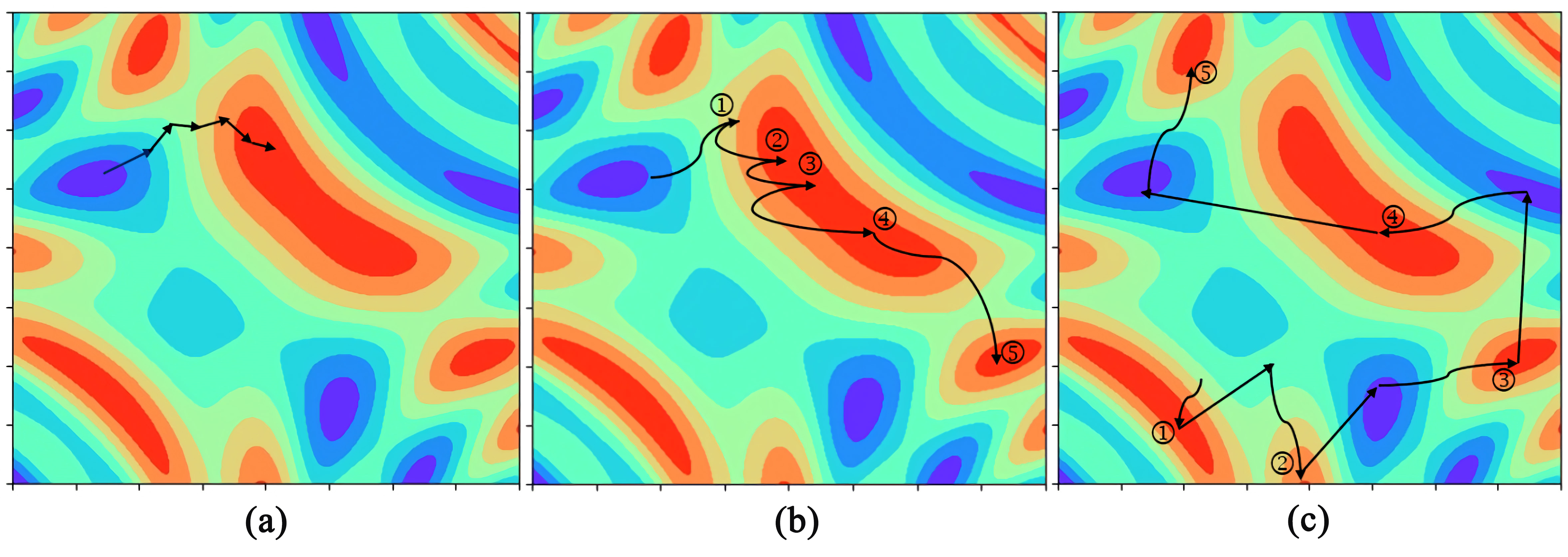}
  \caption{Comparision of different algorithms exploring local optima on the loss surface. (a) Tranditional SGD. (b) Snapshot. (c) Auto-Ensemble}
  \label{fig:1}      
\end{figure*}

\section{Related Work}
\label{sec:1}
Classification ensemble learning techniques have demonstrated powerful capacities to improve upon the classification accuracy of a base learning algorithm \cite{b6}.  A common feature of these approaches is to obtain multiple classifiers by the repeated application of basic learning algorithms to training data. To classify a new sample, we need to obtain the classification results of each classifier, and then aggregate the voting results to get a final classification, this typically achieving significantly better performance than an individual learner \cite{b10}. In some scenarios, the ensemble of simple models can achieve comparable results with complex models, which greatly reduces the computational cost.

Training deep neural network is a process of training loss function composed of a corpus of feature vectors and accompanying labels. Li et al. \cite{b11} proposed a visualization method of deep neural network loss surface, which found that the more complex the network, the more chaotic the loss surface is. And it is also found that there are many local optimal solutions in the large loss surface \cite{b11}\cite{b12}. Ensemble learning makes fully use of these different local minima \cite{b13}.

By scheduling the learning rate, model can converge to different optimal solution. Once the model encounters a saddle point during training, it can quickly jump out of it by increasing the learning rate. The related work of cyclic learning rate(CLR) proves that the CLR schedule can make the convolutional neural network model training process more efficient. And it eliminates the need to perform numerous experiments to find the best values and schedule \cite{b14}.

Research shows that by gathering outputs of neural networks from different epochs at the end of training can stabilize final predictions \cite{b15}. Checkpoint ensemble provides a method to collect abundant models within one single training process \cite{b16}. It greatly shortens training time that ensemble requires and achieves better improvements than traditional ensembles.

Snapshot Ensemble combined the cyclic learning rate with checkpoint ensemble: it adopted a warm restart method\cite{b13}\cite{b17}\cite{b18}, where in each restart the learning rate is initialized to some value and is scheduled to decrease following a cosine function. At each end of cycle, they save the snapshots of model. Multiple snapshot models can be collected in once training, which greatly reduces the training budgets. Wen et al.\cite{b18} proposed a new Snapshot Ensemble method and a log linear learning rate test method. It combines Snapshot Ensemble with appropriate learning rate range, which outperforms the original methods\cite{b13}\cite{b14}. Fast Geometric Ensembling (FGE)\cite{b19} proposed a method to collect models quickly, which finds paths between two local optima, such that the train loss and test error remain low values along these paths. A smaller cycle length and a simpler learning rate curve can be used to collect a model with high accuracy and diversity along the learning curve.

For the ensemble accuracy depends on the number, diversity, and accuracy of individual models, adaptive ensemble aims to find an optimum condition for ensemble learning. Inoue\cite{b20} proposed an early-exit condition based on confidence level for ensemble. It automatically selects the number of ensembled models, which reduce the computation cost. This inspires us to collect models automatically during training.

According to Bengio \cite{b9}, by scheduling the learning rate, the model can find as many different local optimal solutions as possible when exploring the loss surface. Auto-Ensemble method refers to cyclic learning rate schedule. Every time the checkpoint of model is collected, the learning rate rises to escape the local optimal solution. Finally, all the checkpoints of model are used for ensemble to improve the generalization performance of classification. The training epochs, the range of learning rate and the number of training models are adaptive and unpredictable.

Ju et al.\cite{b21} compared relative performance of various ensemble methods, they found that a special ensemble method: Super Learner, achieved best performance among all the ensemble methods. It is cross-validation based ensemble method, and it uses the validation set of the neural networks for computing the weights of Super Learner. Our AE method refers to its idea and proposes a weighted average method, which helps to improve the prediction.

\section{Auto-Ensemble}
Auto-Ensemble(AE) proposed in this paper is scheduling learning rate to control the process of model exploring loss space. After having collected a checkpoint of model, the learning rate rises to escape from it, and start a new searching process. Finally, all the collected model checkpoints are used for ensemble. Our method can explore as many models as possible with high enough accuracy and diversity.

\subsection{Description of Model Diversity}

The main thing is to solve the problem of model diversity. Huang et al.\cite{b13} have discussed the correlation of collected snapshots, and reasonably chose the snapshot models for combination. Auto-Ensemble ensures the diversity of collected checkpoints: Adaptive learning rate schedule can automatically find the local optimal solution by scheduling the learning rate during training process. After having collected a checkpoint of model, by steeply increasing the learning rate, it can automatically jump out of the local optimal solution, then continue to search for other optimal solutions.  Fig. \ref{fig:1} shows the convergence process using different learning rate schedule. Fig. \ref{fig:1}(a) is the convergence process of traditional SGD, which is slow and inefficient to find a local optimal solution. Fig. \ref{fig:1}(b) shows the procedure of collecting snapshots of SnapShot Ensemble(SSE), it enables model to converge to relatively different local minima in loss surface. In Fig. \ref{fig:1}(c) the arrow line indicates the convergence process of Auto-Ensemble. It can be seen that model escapes sharply from current local optimal solution, and then converges to another one that is different from the previous.


\subsection{Metric of Model Diversity}
\begin{figure}[!htbp]
  \centering
  \includegraphics[width=0.5\textwidth]{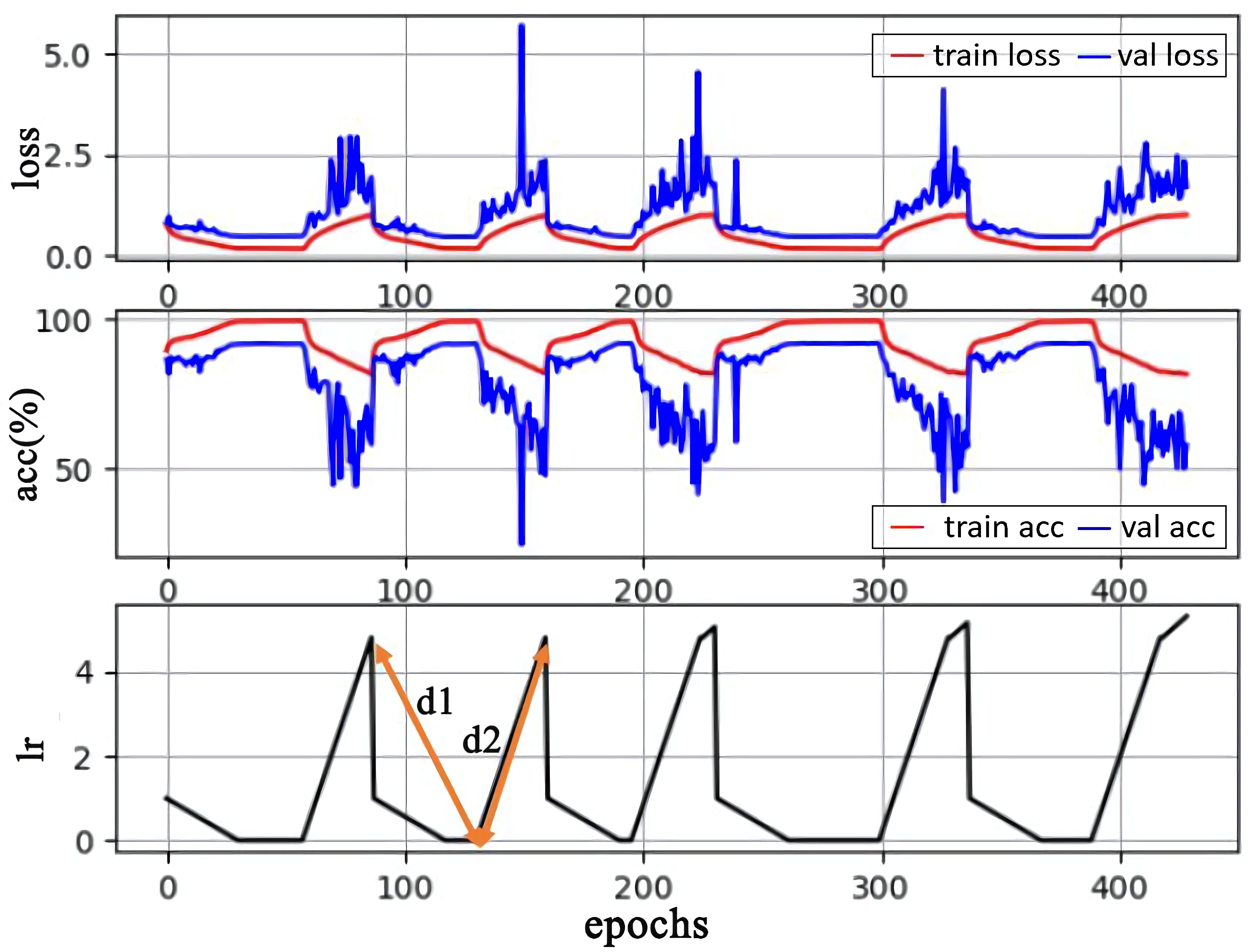}
  \caption{Auto adaptive learning rate schedule.The orange arrow line show two measuring distances during search of loss surfaces}
  \label{fig:2}       
\end{figure}
We found that the weights and biases of the last dense layer (the last but one dense layer of Siamese network) can be extracted to measure the distance between models. We record two Euclidean distances $d_1$ and $d_2$, where $d_1$ is the distance between the weight when model converges to a local optimal solution and the weight when the learning rate rises to the highest in the previous cycle, $d_2$ is the distance between the weights of the checkpoint at the local optimal solution and the weights when the learning rate rises to the maximum in current cycle. In the bottom picture in Fig. \ref{fig:2} the arrows show two measuring distances during searching loss surface. To ensure the distance among collected checkpoints, $d_2$ should be much greater than $d_1$.

\begin{figure*}[h]
  \centering
  \includegraphics[width=\textwidth]{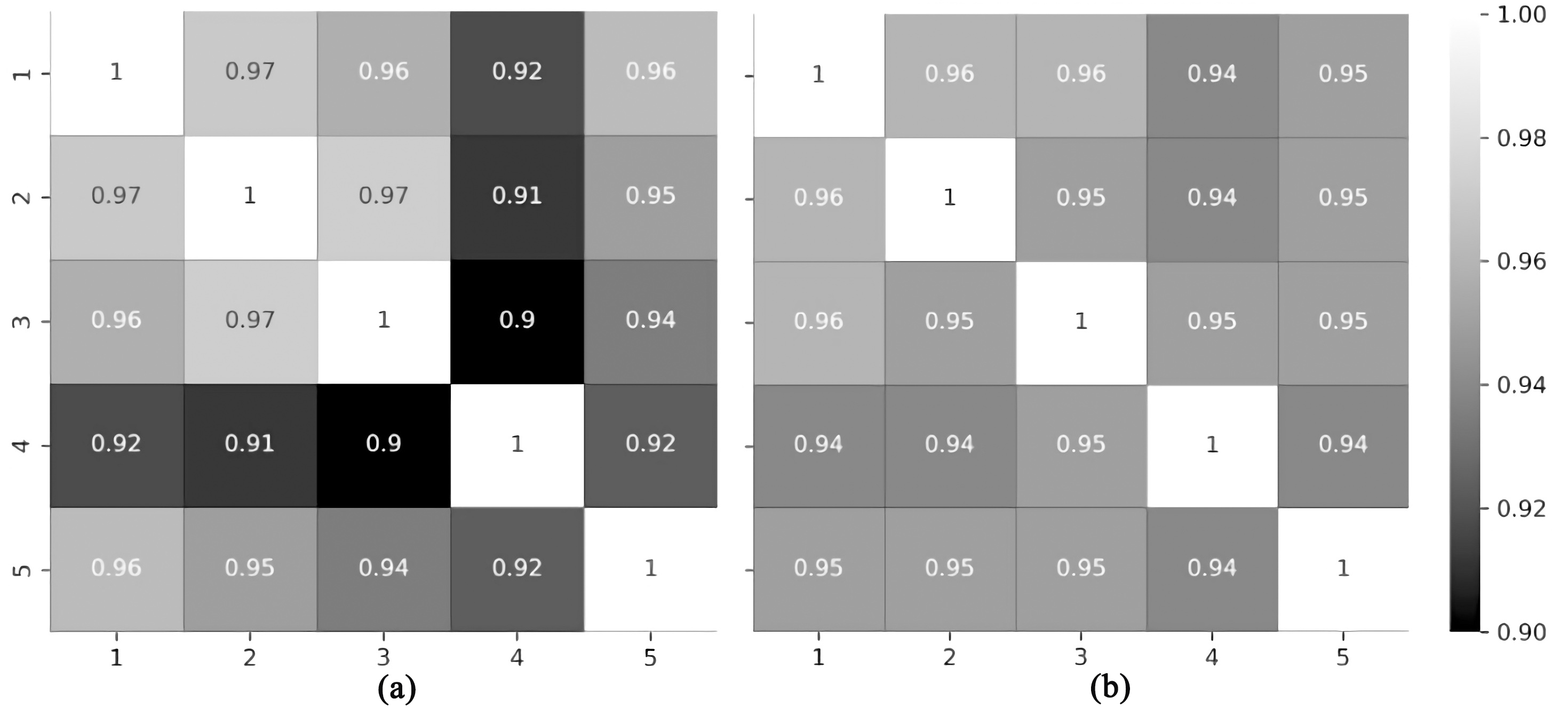}
  \caption{The comparison of distance between weights and correlation coefficient of resnet110 models. (a) The distance among models last dense layer. (b) The correlation coefficient among models.}
  \label{fig:3}       
\end{figure*}

We compared the Euclidean distance between the weights of models with the traditional correlation coefficient method on ResNet models. Fig. \ref{fig:3}(a) shows the distance among models’ last dense layer: to make the comparison more intuitive, we normalized the distance among models, mapping distance 0 to 1 and the maximum distance to 0.9 (corresponding to the maximum and minimum value of the coordinate axis). The distance value y is converted according to the following formula:
\begin{equation}
y =\frac{ \left(y_{max}-y_{min}\right) \left( x-x_{max}\right)}{\left(x_{max}-x_{min}\right)} + y_{min}.
\end{equation}
where $y_{max}$ and $y_{min}$ are the coordinate axis boundary of correlation coefficient(Fig. \ref{fig:3}(b)), x is the actual distance between the weights and $x_{max}$, $x_{min}$ are its maximum and minimum, respectively. Fig. \ref{fig:3}(b) is the correlation coefficient among models. It is observed that the farther the model is, the greater the distance (the smaller the normalized distance value) and the smaller the correlation, preliminary proved that the distance between weights can be used to measure the diversity of models.

\subsection{Learning Rate Schedule}
The learning rate schedule adopts piecewise linear cyclic learning rate schedule refers to Garipov et al.\cite{b19}. We set the appropriate learning rate boundaries $\alpha_1$ and $\alpha_2$($\alpha_1>\alpha_2$) for each neural network(the method of setting learning rate boundary will be introduced in Chapter 4). The values of $\alpha_1$ and $\alpha_2$ are quite different (generally two orders of magnitude), where $\alpha_1$ is to speed up the gradient descent process, while $\alpha_2$ is to make the model converge to a wide local optimal solution. In the cycle of collecting one checkpoint, the learning rate decreases linearly with the change rate $\beta$, and then remains at the minimum until model converges. Formally, the learning rate \emph{lr} has the form:
\begin{equation}
lr=\begin{cases}
\alpha_1-\beta n & n\leq N\\
\alpha_2 & n>N\\
\end{cases}
\end{equation}
where \emph{n} is the total number of training iterations, change rate $\beta=(\alpha_1-\alpha_2)/N$. Then learning rate increases linearly. Learning rate rise phase is divided into two parts: rapid rise phase and loss surface exploring phase. The learning rate \emph{lr} has the form:
\begin{equation}
lr=\left\{ \begin{matrix}
   {{\beta }_{1}}(n-M)+{{\alpha }_{2}} & n\text{ in upward phase \uppercase\expandafter{\romannumeral1} }  \\
   {{\beta }_{2}}(n-M-m)+l{{r}_{now}} & n\text{ in upward phase \uppercase\expandafter{\romannumeral2} }  \\
\end{matrix} \right.
\end{equation}
where M is the total number of training iterations from the beginning till now. \emph{m} is the length of the rapid rise phase. $lr_{now}$ is the learning rate in the end of the rapid rise phase. The change rate of learning rate($\beta_1$) in the rapid rise phase is the largest, which aims to make the model jump out of the current local optima quickly. Then the learning rate increases slowly with the change rate $\beta_2$, aims to explore the loss surface more carefully.
The change rate of learning rate subsequently($\beta_2$) declines to explore the loss surface more carefully.
\subsection{Auto-Ensemble}
\begin{algorithm}[t]
\caption{Auto-Ensemble Algorithm}
\label{alg:A}
\begin{algorithmic}
\REQUIRE ~~\\
LR bounds $\alpha_1$,$\alpha_2$, LR change rate $\beta \text{=}\frac{({{\alpha }_{1}}-{{\alpha }_{2}})}{N}$\\
${{\beta }_{1}}\text{=}\frac{({{\alpha }_{1}}-{{\alpha }_{2}})}{aN}$, ${{\beta }_{2}}\text{=}\frac{({{\alpha }_{1}}-{{\alpha }_{2}})}{\text{b}N}$ $(a>b>1)$\\
number of iterations \emph{n}, epochs of rapid rise phase \emph{m}, ratio of $d_2$ to $d_1$:$\alpha$\\
\textbf{Pretrain phase:}adopt 75\% of epochs to run standard learning rate schedule.\\
\ENSURE ~~\\
\REPEAT
\REPEAT
\STATE $lr={{\alpha }_{1}}-\beta n$
\IF{$n > N$ and the model has not converged}
\STATE $lr={{\alpha }_{2}}$
\ENDIF
\UNTIL{the model is converged and then collect the checkpoint, record number of iterations \emph{M}}

\REPEAT
\FOR{\emph{n} in \emph{M}+\emph{m}(rapid rise phase)}
\STATE $lr=\beta {}_{1}(n-M)+{{\alpha }_{2}}$
\ENDFOR, record the current learning rate $lr_{now}$
\STATE $lr={{\beta }_{2}}(n-M-m)+l{{r}_{now}}$
\UNTIL{${{d}_{2}}>\alpha *{{d}_{1}}$, the current cycle is over}
\UNTIL{satisfy the conditions for training stopping}
\STATE \textbf{Ensemble phase:}
\\For each model:${{\theta }_{0}}\cdots \theta {}_{\text{T}}$, get the predicted softmax output ${{h}_{\theta }}(x)$, where \emph{T} is the total number of collected checkpoints, \emph{x} is the training data.
\\Define a fully collected network H(x) to train the weight:\\
\begin{center}
    $H(x)=f(concat[{{h}_{{{\theta }_{1}}}}(x)\cdots {{h}_{{{\theta }_{T}}}}(x)])$
\end{center}
The weighted averaging result is:\\
\begin{center}
    $H(x)=\sum\nolimits_{i=1}^{T}{{{w}_{i}}{{h}_{{{\theta }_{i}}}}(x)}$
\end{center}
where $w_i$ is the weight of individual learner $\theta_i$
\end{algorithmic}
\end{algorithm}
\noindent The procedure is summarized in Algorithm \ref{alg:A}. Before starting to schedule the learning rate, we adopt a pretrain phase. Warm start plays a key role in machine learning: before training a model, if the model is pretrained for a period, the experimental results tend to be significantly improved. After having collected several model checkpoints, the ensemble prediction at test time is the average of every model’s softmax outputs. In addition, in order to improve the efficiency of ensemble, we have designed a weighted averaging method, each model is weighted with different weights.

Simple averaging means to average the softmax output of each model, while weighted averaging gives weights to the output of each model.The weight is generally learned from the validation set, which is generated from the training set through data augmentation, etc. Then we use test set to get the test accuracy. Our algorithm designs an one layer fully connected network to learn such weights automatically. It's worth mentioning that the bias of fully connected network is fixed to ZERO. The weight is initialized as a one-dimensional vector with length \textit{T}, where \textit{T} is the number of collected checkpoints. When training fully connected network, the training data is the 1D array and the label is the original label. After training, the weight of layer is used to ensemble checkpoints.

The above method can improve the ensemble accuracy obviously and smooth the uneven distribution of model accuracy.

\section{Experiments}
We demonstrate the effectiveness of Auto-Ensemble on different datasets and networks, we compare our method with related state-of-the-art. And run all experiments with Keras.
\subsection{Dataset}
\setcounter{subsubsection}{1}
\paragraph{CIFAR} The CIFAR-10 and CIFAR-100 are labeled subsets of the 80 million tiny images dataset \cite{b22}. The CIFAR-10 dataset consists of 60000 32x32 colour images in 10 classes, with 6000 images per class. There are 50000 training images and 10000 test images. We used a standard data augmentation scheme in Keras document. The augmentation scheme is different in VGG, compared with ResNet and Wide ResNet.

\paragraph{Omniglot} The Omniglot dataset is collected by Brenden Lake and his collaborators at MIT via Amazon’s Mechanical Turk to produce a standard benchmark for learning from few examples in the handwritten character recognition domain\cite{b23}. The Omniglot dataset consists of 1,623 handwritten characters, with only 20 samples per class. The dataset is divided into training set and test set: 964 classes for train and 659 classes for test.

\subsection{Training Setting}
\subsubsection{Architecture} We test several classic neural networks: including residual networks(ResNet)\cite{b24}, Wide ResNet\cite{b25} and VGG16\cite{b26}. For ResNet, we use the original 110-layer network(ResNet110), for Wide ResNet we use a 28-layer Wide ResNet with widening factor 10(WRN-28-10). We use the same standard data augmentation scheme on CIFAR10 and CIFAR100.

\subsubsection{Hyperparameters} Our experiments used following hyperparameters: For WRN-28-10 we set the learning rates: ${{\alpha }_{1}}\text{=}0.5$, ${{\alpha }_{2}}\text{=}0.5\times {{10}^{-3}}$; for ResNet110 ${{\alpha }_{1}}\text{=}0.5$, ${{\alpha }_{2}}\text{=}0.01$; for VGG16 ${{\alpha }_{1}}\text{=}0.4$, ${{\alpha }_{2}}\text{=}0.01$. For few-shot learning the learning rate boundary is 0.005-0.03. For distance measurement, $\alpha$ was set to 1.5 for VGG16, WRN-28-10 and ResNet110. For few-shot learning $\alpha$ was set to 2. For pretraining epochs, Resnet110, WRN-28-10 and VGG16 have 110, 45 and 100 pretrain epochs respectively. In few-shot learning, the pre-training phase contains 45000 tasks. We will discuss the effects of these parameters on the experimental results in the following sections.

\begin{figure*}[!htbp]
  \centering
  \includegraphics[width=\textwidth]{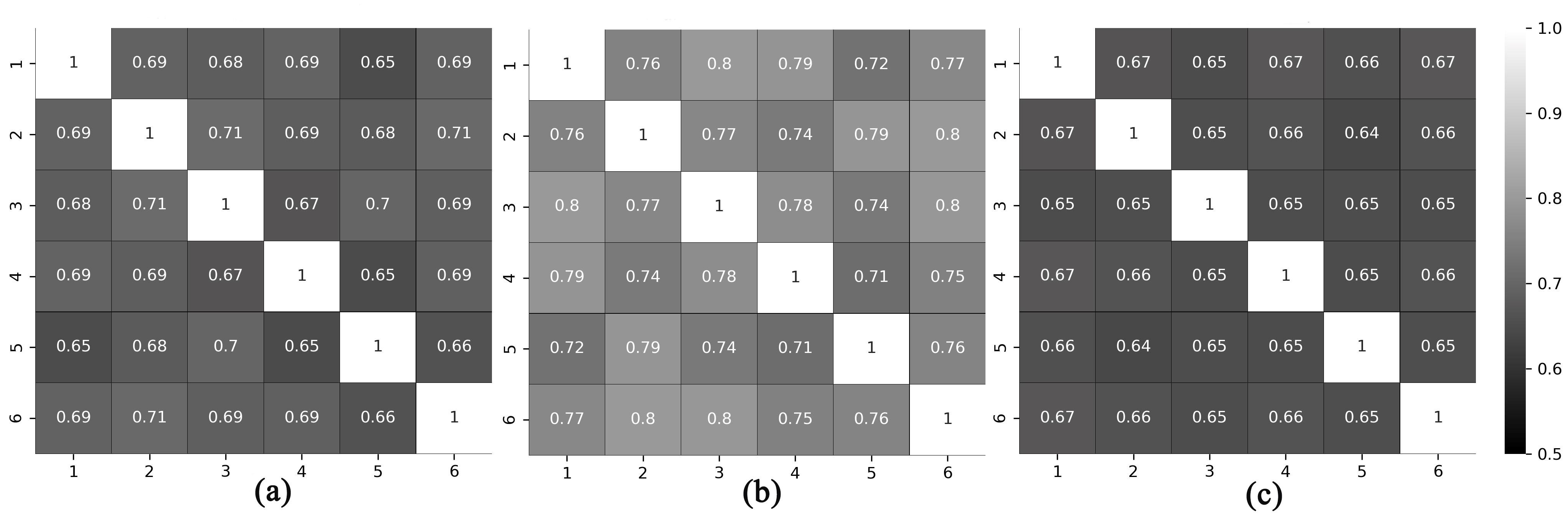}
  \caption{Pairwise correlation of softmax outputs between two models collected by AE, SSE and RIE. (a) Auto Ensemble(AE) with adaptive learning rate cycle. (b) Snapshot, with cosine annealing cycles(restart with ${{\alpha }_{0}}\text{=}0.1$ every 40 epochs). (c) Random initialization ensemble models(RIE) with traditional learning rate schedulers}
  \label{fig:4}       
\end{figure*}

\subsection{Baseline and Comparison}
\subsubsection{Comparison in Model Collection}Auto-Ensemble has a unique learning rate scheduling method to collect models. To illustrate our advantage, different learning rate schedulers are implemented. We collected checkpoint ensemble models(CE) and random initialization ensemble models(RIE) with traditional learning rate schedulers\cite{b16}. Besides, Cosine Cyclical Learning Rate scheduler (SSE or CCLR)\cite{b13}, Max-Min Cosine CLR (MMCCLR)\cite{b18} and Triangular CLR(FGE)\cite{b19} are also used for comparison. The baseline model is a single independently trained network (Ind) using stepwise decay SGD.

It is worth mentioning that for fair comparing with state-of-the-art, we reimplement above methods, using the same architectures and data augmentations. All parameters of them follow those in paper when we reproduce them. For CE and RIE, we reimplement these methods based on our datasets and models. For SSE, we totally reproduce experiment according to its paper. We didn’t reproduce MMCCLR for its learning rate scheduler is similar to SSE. For FGE, we only refer to its Triangular CLR as stated in its paper(its implements of curve finding experiments changes the way of gradient update of model to some extent).

\subsubsection{Comparison in Ensemble Method}After having collected ensemble models, there have been some rules to combine them together to make predictions. We reimplemented Adaptive Ensembles based Confidence Intervals(CIs)\cite{b20}, which automatically select number of models for ensemble: we used the confidence-level based early-exit condition with a 95\% confidence level for all datasets.

And the weighted averaging method are compared with Super Learner(SL). We referred to Ju et al.\cite{b21} and directly extracted the experimental result from them. The results show that our weighted averaging method can make the ensemble more robust.

\subsection{Auto-Ensemble on Supervised Learning}
\subsubsection{Ensemble Result}

\begin{table*}[!htbp]
\centering
\caption{Ensemble Accuracys and Improved Accuracys(\%) on CIFAR-10 and CIFAR-100 datasets}
\label{tab:1}
\begin{tabular}{ccccccccccccc}
\hline
         & \multicolumn{4}{c}{ResNet110}                         & \multicolumn{4}{c}{WRN-28-10}                         & \multicolumn{4}{c}{VGG16}                             \\ \hline
Method   & \multicolumn{2}{c}{C10}   & \multicolumn{2}{c}{C100}  & \multicolumn{2}{c}{C10}   & \multicolumn{2}{c}{C100}  & \multicolumn{2}{c}{C10}   & \multicolumn{2}{c}{C100}  \\ \hline
Ind      & \multicolumn{2}{c}{\textbf{94.18}} & \multicolumn{2}{c}{73.21} & \multicolumn{2}{c}{94.28} & \multicolumn{2}{c}{74.74} & \multicolumn{2}{c}{92.92} & \multicolumn{2}{c}{69.35} \\
RIE      & 94.51        & 0.33       & 74.36        & 1.15       & 95.61        & 1.33       & 77.47        & 2.73       & \textbf{93.7}         & \textbf{0.78}       & 70.29        & 0.94       \\
SSE      & 94.03        & 0          & 71.59        & 0          & 95.55        & 1.27       & 76.38        & 1.64       & 93.19        & 0.27       & 70.28        & 0.93       \\
FGE      & 93.74        & 0          & 72.43        & 0          & 94.21        & 0          & 76.12        & 1.38       & 93.26        & 0.34       & 69.53        & 0.18       \\
CE       & 94.11        & 0          & 73.92        & 0.71       & 95.05        & 0.77       & 77.17        & 2.43       & 92.8         & 0          & 71.15        & 1.8        \\
CIs      & 94.37        & 0.19       & 74.46        & 1.25       & 95.22        & 0.94       & 77.41        & 2.67       & 93.63        & 0.71       & 70.56        & 1.21       \\
AE       & \textbf{94.87}        & \textbf{0.69}       & \textbf{76.55}       & \textbf{3.34}       & \textbf{95.91}        & \textbf{1.63}       & \textbf{77.71}        & \textbf{2.97}       & 93.61        & 0.69       & \textbf{71.29}        & \textbf{1.94}       \\
AE(full) & 95.05        & 0.87       & 77.18        & 3.97       & 96.09        & 1.81       & 79.26        & 4.52       & 93.93        & 1.01       & 72.16        & 2.81       \\ \hline
\end{tabular}
\end{table*}

All the results are summarized in TABLE \ref{tab:1}. For each method, we used simple averaging methods to ensemble models. And we list the improved accuracy compared to single model(Ind). For AE we show the weighted averaging result. These results are obtained by a fully connected network trained by validation set. We set the learning rate at the range of 0.01-0.001 and select appropriate learning rate that maximizes ensemble accuracy.

Our Auto-Ensemble (AE) results were compared with SSE, FGE, CE, RIE, Adaptive Ensemble based Confidence Intervals (CIs) and independently trained networks (Ind). The best ensemble results are \textbf{bolded} in TABLE \ref{tab:1}. Experiment showed that in most cases, the ensemble accuracys of Auto-Ensemble were better than other methods. And compared with single model, the improved accuracy is considerable. SSE, FGE and CE sometimes have no improvement because of the poor diversity.

\begin{table}[bt]
	\centering
	\caption{ResNet110 Checkpoint Ensemble (CE) performance on CIFAR-10 for varying ensemble method.}
	\label{tab:2}
	\begin{tabular}{cc}
		\hline
		Ensemble method                               & Ensemble Result (\%) \\ \hline
		Unweighted Average                            & 93.54                 \\
		Super Learner                                 & 88.0                   \\ \hline
		Distribution of Collected Models              & $91.42\pm1.98$          \\ \hline
	\end{tabular}
\end{table}

We also add several Super Learner(SL) results to compare the effectiveness of our ensemble method. The results are directly extracted from the original paper(TABLE \ref{tab:2}). Clearly, our weighted averaging method takes advantage of SL and achieves best combination.

\subsubsection{Diversity of Models}

To illustrate the higher diversity of our models, we calculated the correlation of the softmax output of each pair of models. Fig. \ref{fig:4} shows the correlation coefficients among models collected by different methods: Fig. \ref{fig:4}(a) stands for Auto-Ensemble(AE) with adaptive learning rate cycles. Fig. \ref{fig:4}(b) shows Snapshot with cosine annealing cycles. Fig. \ref{fig:4}(c) is RIE with independently trained networks. The correlation of Snapshot is higher than that of the other two methods, indicating that there is less diversity between models. Although AE’s diversity of models is not as good as RIE, compared with RIE, it reduces training budget and can collect models with enough diversity.

\subsubsection{Training Budget}
Our Auto-Ensemble has unfixed time and storage complexity: storage complexity depends on the diversity of collected model. As stated in our paper, there is a hyperparameter $\alpha$ to adjust the distance among models. The number of collected models can’t be specified because it is related to the value of $\alpha$. The training budget is adaptive for AE: the distance among models is adjustable, and the farther the model is, the more epochs needed to collect one model.

While these are fixed for SSE and FGE: the number of epochs can be specified to collect the target model. In our experiment, the ensemble size of SSE is 5: we ensemble the last 5 snapshots in once training. For CE the storage complexity is large: at each epoch, we save the checkpoint for later ensemble. For RIE, it requires to run a single model separately several times: according to Chen, Lundberg, and Lee\cite{b16}, they ensembled 5 models. The time complexity is the largest. For adaptive ensemble, we select among RIE models, the time and storage complexities are less than RIE.

\begin{table*}[bt]
\centering
\caption{Computational Expenses on ResNet with CIFAR-10}
\label{tab:3}
\begin{tabular}{ccc}
\hline
Ensemble Method & Average Epochs per Ensemble Model & Ensemble Size \\ \hline
AE              & 62                                & 12            \\
SSE             & 40-80                             & 5             \\
FGE             & 23                                & 6             \\
CE              & 1-3                               & 68-150        \\
RIE             & 200                               & 5             \\
CIs             & 200                               & 5             \\
Ind             & 200                               & 1             \\ \hline
\end{tabular}
\end{table*}

Table \ref{tab:3} shows the average number of epochs required to train a ResNet110 model on CIFAR10. The comparison of computational expense is almost the same on other models and dataset.

Above all, the order of ensemble size is: $SSE\approx CIs<RIE<AE<FGE<CE$, the order of time budget is: $Ind=CE=SSE<FGE<AE<CIs<RIE$. It is worth mentioning that the average epoch of SSE is unfixed: the cycle is 40 but we collect last 5 models for better convergence, so the number of average epochs is more than 40 and reaches 80(collect 10 snapshots and ensemble the last 5). For CE, we select the first M best models.

\subsection{Study of Auto-Ensemble}
\subsubsection{Learning Rate Boundary}It’s necessary to set different learning rate boundaries for different neural networks in the decline phase, which is to make the model converge better. However, the learning rate is not limited during the rise phase. Refer to Smith\cite{b14}, this paper provides a method to determine learning rate boundaries as following steps:

\begin{figure}[tb]
  \centering
  \includegraphics[width=0.5\textwidth]{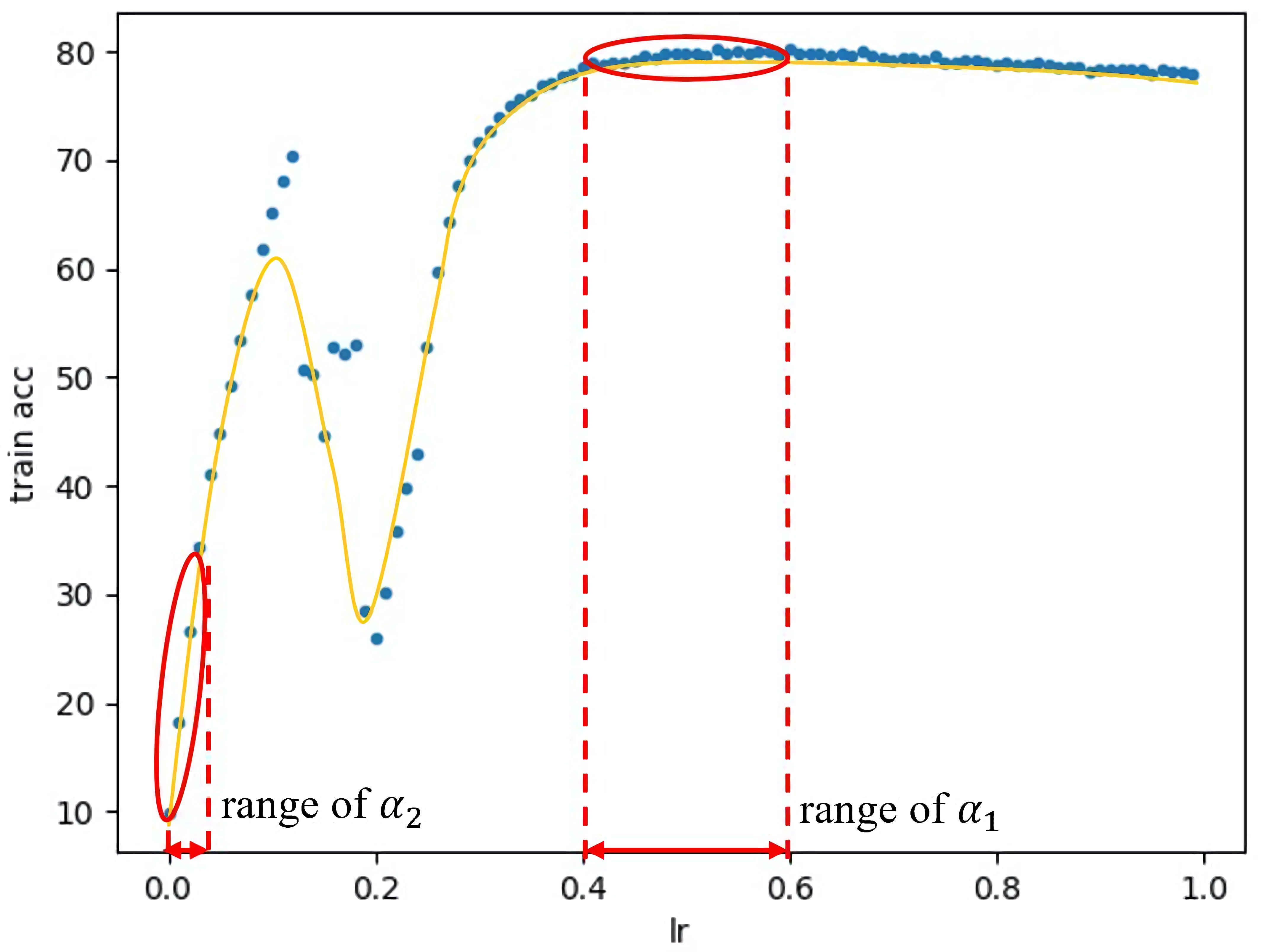}
  \caption{Accuracy-LearningRate curve which used to estimate the rough range of cyclic learning rate boundaries.}
  \label{fig:5}       
\end{figure}

1) Pre-train the certain neural network for a period of time, let the learning rate increase linearly within a rough boundary(usually 0.0001-1).

2) Fit the training accuracy curve changing with the learning rate according to the training accuracy in the training process.

3) Determine the learning rate boundaries $\alpha_1$ and $\alpha_2$ through the Accuracy-LearningRate curve. The range of learning rate when the training accuracy improved significantly at the beginning stage is used as the candidate range of lower boundary $\alpha_2$. Select the upper boundary $\alpha_1$ from the learning rate range where the Accuracy-LearningRate curve becomes flatten. As shown in Fig. \ref{fig:5}, this method can only provide a rough scope of $\alpha_1$ and $\alpha_2$.

\begin{table*}[tb]
\centering
\caption{VGG16 Auto-Ensemble performance on CIFAR-10 for varying learning rate boundary.}
\label{tab:4}
\begin{tabular}{ccc}
\hline
Learning Rate Boundary & Ensemble Result (\%) & Distribution of Collected Models (\%) \\\hline
0.5-$0.5*10^{-2}$           & 93                   & $91.95\pm0.4$                             \\
0.4-0.01               & 93.68                & $92.71\pm0.22$                            \\
0.6-0.01               & 93.08                & $92.12\pm0.31$                            \\ \hline
\end{tabular}
\end{table*}

Fig. \ref{fig:5} provides the Accuracy-LearningRate curve obtained by the pre-training of VGG16 on CIFAR10. The training accuracy rises rapidly at the beginning of training, so it’s reasonable to choose $\alpha_2$ between ${{10}^{-4}}$ and ${{10}^{-2}}$. The accuracy improved slightly after learning rate increased to 0.4, and tended to be saturated when learning rate reached 0.6, so $\alpha_1$ can be selected in the range of 0.4-0.6. Based on the learning rate boundaries provided by Fig. \ref{fig:5}, we carried out the study on the impact of different learning rate boundaries on the performance of Auto-Ensemble, the results are shown in TABLE \ref{tab:4}. We found that those boundaries influence the ensemble performance slightly. However, it’s necessary to choose learning rate boundaries carefully within above ranges to get better performance of ensemble models.

In addition, we need to set the change rate of learning rate. The whole learning rate schedule is divided into three phases: the decline phase, the rise phase 1 and the rise phase 2. In the decline phase, the change rate of learning rate is ${({{\alpha }_{1}}-{{\alpha }_{2}})}/{N}\;$. N is usually set to $20\sim30$, in the decline phase $N=25$, in the rise phase $N=5$.We set the learning rate change rate in the decline phase, the rise phase 1 and the rise phase 2 separately to $\beta$, $\beta_1$, $\beta_2$, then $\beta \approx {{\beta }_{1}}<{{\beta }_{2}}$.


\subsubsection{Pretrain}
\begin{table}[tb]
\centering
\caption{ VGG16 Auto-Ensemble performance on CIFAR-10 for varying pretrain epochs}
\label{tab:5}
\begin{tabular}{p{40pt}p{50pt}p{40pt}}
\hline
Pretrain Epochs(rounds) & Best Model Accuracy(\%) & Ensemble Result(\%) \\ \hline
0                       & 92.08                   & 92.31               \\
40                      & 92.75                   & 93                  \\
80                      & 92.49                   & 93.21               \\
120                     & 92.71                   & 92.43               \\ \hline
\end{tabular}
\end{table}

It is noteworthy that the effect of the pre-training epochs on the collection of the model: If the pre-training model is under-fitting, the difficulty of collecting the first convergent checkpoint of model will increase, which will affect the subsequent ensemble work. If we start from an over-fitting model, it will also lead to poor accuracy and diversity of the collected models. TABLE \ref{tab:5} shows the effect of different pre-training epochs on ensemble accuracy of VGG models. It can be seen that 80 is an appropriate pre-training epoch.

\subsubsection{Parameter of Diversity}
\begin{table}[tb]
\centering
\caption{ Distribution of Model Accuracy with different $\alpha$}
\label{tab:6}
\begin{tabular}{p{25pt}p{55pt}p{85pt}}
\hline
Value of $\alpha$ & Ensemble Result (\%) & Distribution of Collected Models (\%) \\ \hline
1.2 & 93.16 & $92.14\pm0.87$ \\
1.4 & 93.02 & $92.24\pm0.44$ \\
1.6 & 93 & $91.96\pm0.4$ \\
1.8 & 93.16 & $92.29\pm0.37$ \\ \hline
\end{tabular}
\end{table}

The learning rate stops increasing when ${{d}_{2}}>\alpha *{{d}_{1}}$. TABLE \ref{tab:6} shows the effect of different value of $\alpha$ on the collection of VGG models. Experiments show that the model is not sensitive to the value of $\alpha$. But if the distance is too large, it will cause the model to escape too far to converge, the loss and accuracy also become unpredictable. So we limit $1<\alpha<2$.

\subsubsection{Conditions for Stop Training}
The condition for the experiment to stop is a question worth discussing, because the number of model checkpoints needed to collect in the training process is unknown.
\begin{figure*}[tb]
  \centering
  \includegraphics[width=\textwidth]{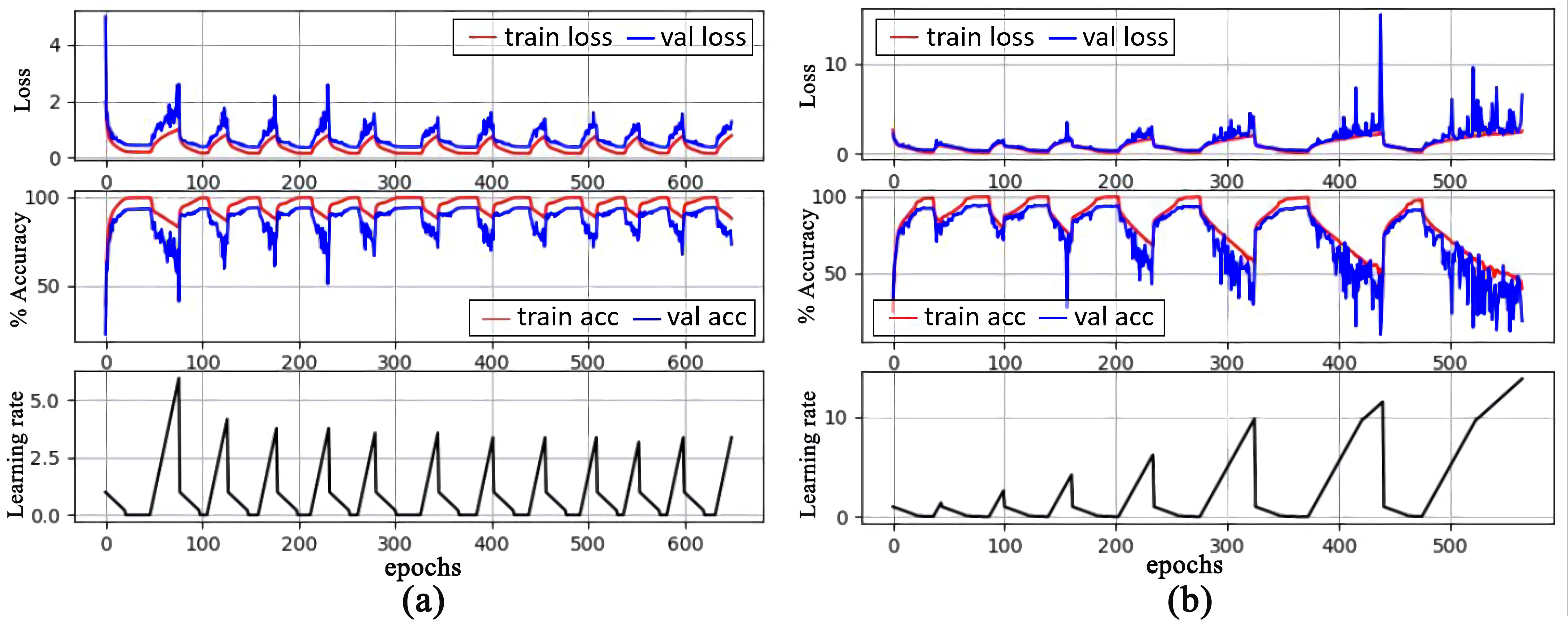}
  \caption{Training process of ResNet and Wide Resnet models.(a) The training process of ResNet110. (b) The training process of WRN-28-10.}
  \label{fig:6}       
\end{figure*}

\begin{figure*}[bt]
	\centering
	\includegraphics[width=\textwidth]{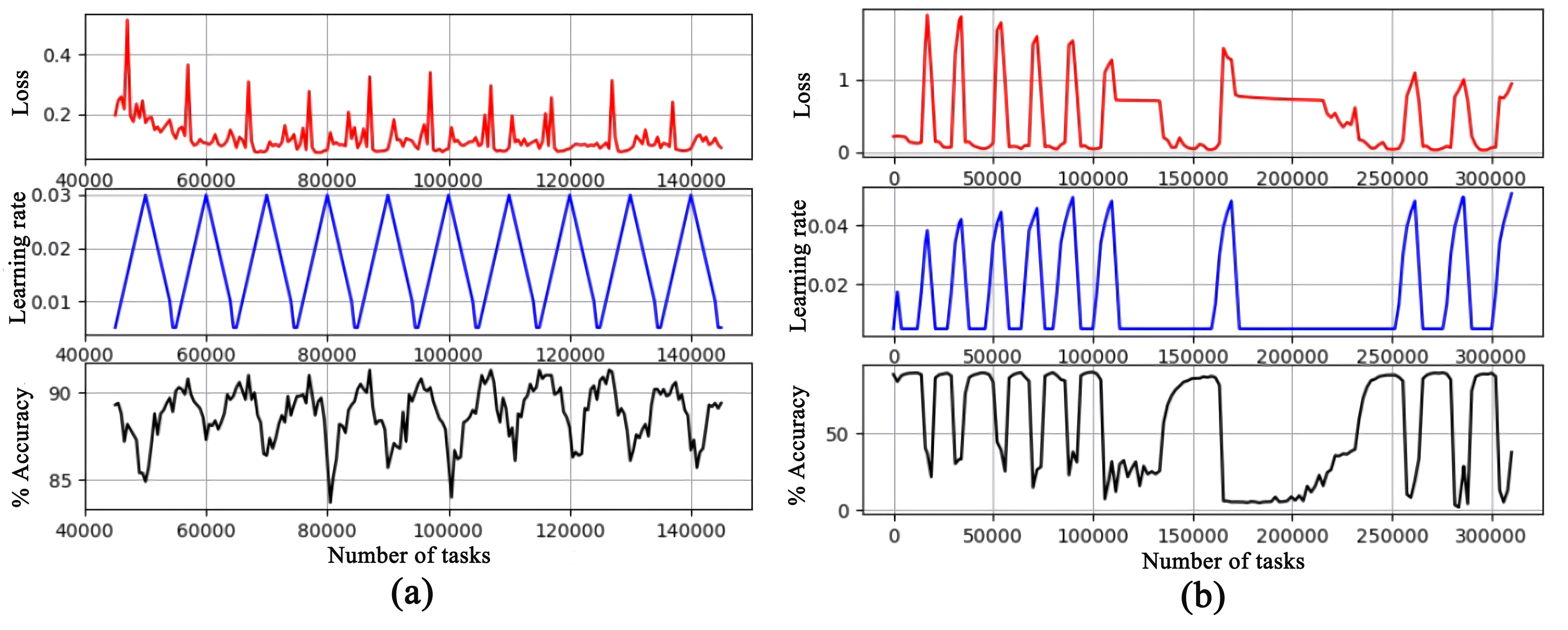}
	\caption{The comparison of piecewise linear cyclic learning rate schedule and auto adaptive learning rate schedule in few-shot learning.(a) Cyclic learning rate schedule. (b) Auto adaptive learning rate schedule.}
	\label{fig:7}       
\end{figure*}

\begin{table}[t]
	\centering
	\caption{Performance of Auto-Ensemble on Siamese Networks}
	\label{tab:7}
	\begin{tabular}{cc}
		\hline
		Model                                  & 20-way 1 shot acc (\%) \\ \hline
		Matching Network \cite{b28} & 93.8                   \\
		Siamese Network \cite{b27}     & 88.0                     \\
		Siamese Network with ensemble method   & 92.2                   \\ \hline
	\end{tabular}
\end{table}

Experiments show that it is easy for VGG16 and ResNet110 to collect models. As long as the training process does not stop, infinite models will be collected. So, the number of checkpoints can be used to limit the conditions for training stopping. Fig. \ref{fig:6}(a) is the training curve of ResNet: Two nearly coincident curves represent the loss and accuracy changes of training set and test sets. The processes of collecting WRN-28-10 and Siamese Network are relatively tough. Fig. \ref{fig:6}(b) demonstrates WRN-28-10 training process: With the increase of training epochs, the learning rate should increase to very high level to escape the local optimal solution, and the accuracy of collected checkpoint becomes much lower. In this case, it can stop training when the learning rate has increased to a certain value, for the accuracy of the checkpoints collected next is not high enough.

Inoue\cite{b20} proposed an early-exit method based on the confidence level of local prediction, we reimplemented it and found that the adaptive ensemble result is not as good as that of ensembling all models(RIE). The result is shown in TABLE \ref{tab:1}

\subsection{Auto-Ensemble on Few-shot Learning}
Our mainly contribution also contains the application in few-shot learning, which indicates that our AE method can apply to some non-traditional supervisory problems. Few-shot learning is not a new research topic, and there have been many state-of-the-art. To verify the effectiveness of Auto-Ensemble, we choose Siamese Network as a baseline for its relatively poor performance. The experimental results show that AE method has brought significant improvement.

We used the Siamese Neural Network \cite{b27}, but added a dense layer before the last neural layer. Since the weight of this dense layer is used to measure the distance between the models during the training process. The learning rate boundary is 0.005-0.03. The distance measurement $\alpha$ is set to 2. The pre-training phase contains 45000 tasks. We experimented on the Omniglot dataset, which is divided into training set and testing set according to Vinyals et al.\cite{b28}.

\begin{figure*}[bt]
	\centering
	\includegraphics[width=\textwidth]{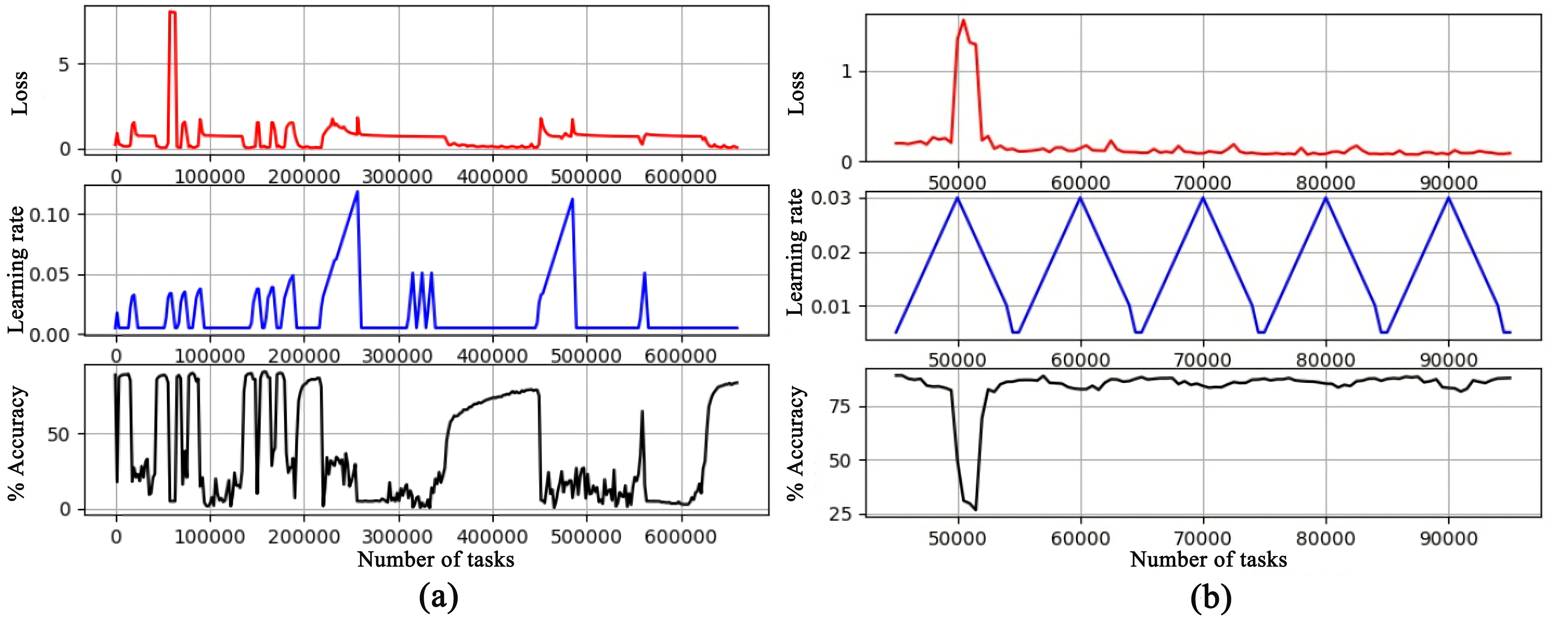}
	\caption{Different situations of two learning rate schedules when model reaches a saddle point. (a) Auto adaptive learning rate schedule. (b) Cyclic learning rate schedule.}
	\label{fig:8}       
\end{figure*}

TABLE \ref{tab:7} illustrates the performance of Auto-Ensemble: The accuracy we obtained with weighted averaging (92.2\%) has great improvement on an original Siamese Network model, and almost catches up with more complex models Matching Networks \cite{b28} (93.8\%). Our experiment collected 10 checkpoints of model.

Fig. \ref{fig:7} illustrates the use of two methods to collect the checkpoints of model: Using the cyclic learning rate schedule with 10000 tasks in one cycle, the accuracy of the model does not change much with the learning rate. Due to the instability of the model, the model often gets stuck in the saddle point during the training process, where the model is difficult to converge. Fig. \ref{fig:8}(b) shows that cyclic learning rate may be invalid when model reaches a saddle point. But in Auto-Ensemble, the learning rate jitters to let the model jump out of the saddle point automatically (Fig. \ref{fig:8}(a)). Adaptive learning rate schedule can fully mine different checkpoints of model.

\section{Conclusions}
This paper proposed an adaptive learning rate schedule for ensemble learning: by scheduling the learning rate, the model can converge and then escape from the local optimal solutions. We pay attention to the improvement of performance rather than the absolute performance. By collecting the checkpoint of models, the ensemble accuracy can greatly exceed accuracy of single model. Besides, we proposed a method to measure the diversity among models, so that we can guarantee the diversity of collected models. In some non-traditional supervise problems, like Few-shot Learning, this method can be used to improve the performance of model simply and quickly. We refer to various related work and compare our method with these methods to analyse the results. To verify the effectiveness of our method, we need more experiment on other networks: DenseNet, and Matching Networks used for few-shot learning. In future work we will focus on the optimization of Auto-Ensemble: how to shorten the unpredictable training process. Since the purpose of the training is to collect as many models as possible, but the time and resources required for training are unpredictable, in order to save computing resources, the training process needs to be simplified. We will also focus on improving the method of measuring diversity among models. We only measured the diversity between two adjacent models,  next we will focus on the unpaired diversity measurement method.

\begin{IEEEbiography}[{\includegraphics[width=1in,height=1.25in,clip,keepaspectratio]{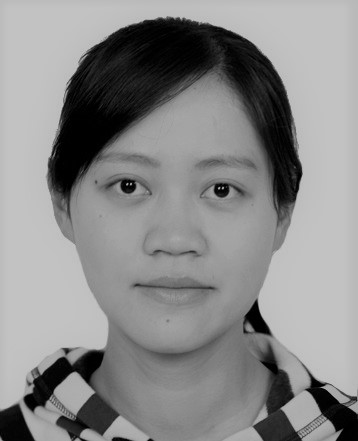}}]{Jun Yang} is studying for a master's degree in school of Electronic Information and Communications(EIC) from Huazhong University of Science and Technology (HUST), Wuhan, China, from 2018 to 2021. Her interests include machine learning, especially few-shot learning and ensemble learning.

\end{IEEEbiography}
\begin{IEEEbiography}[{\includegraphics[width=1in,height=1.25in,clip,keepaspectratio]{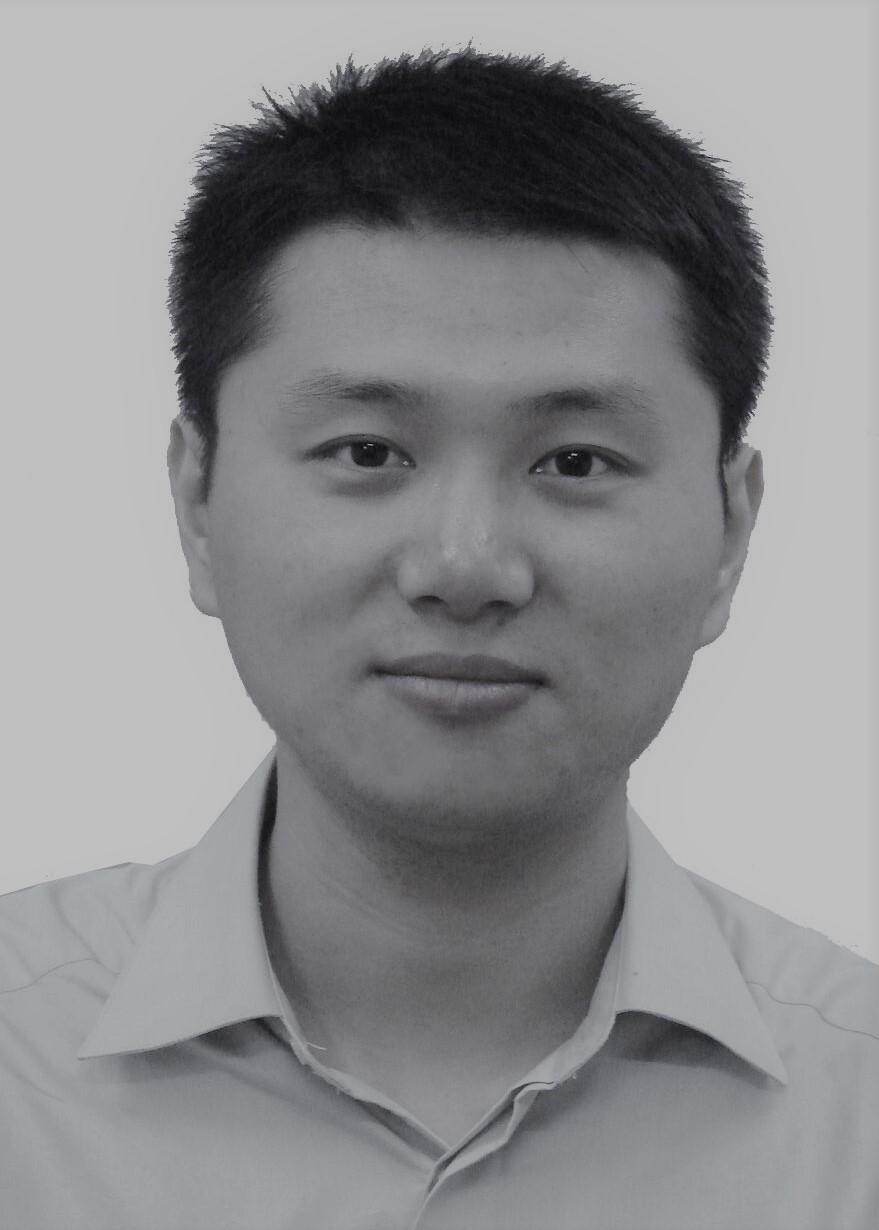}}]{FEI WANG} received the M.S. and PH.D. degrees in school of Electronic Information and Communications(EIC) from Huazhong University of Science and Technology (HUST), Wuhan, China, in 2002 and 2008, respectively. Currently, he is an associate professor at school of EIC, HUST, where he gives courses in probability \& statictis, and stochastic processes. His research work concerns machine learning and its application in smart grid.
\end{IEEEbiography}
\EOD

\begin{thebibliography}{00}



\bibitem{b1}
Blum AL, Rivest RL, ``Original contribution: Training a 3-node neural
  network is np-complete'' 5(1):117--127, 1992.

\bibitem{b2}
Elsken T, Metzen JH, Hutter F,``Neural architecture search: {A} survey.'' J
  Mach Learn Res 20:55:1--55:21, 2019.

\bibitem{b3}
Fang J, Chen Y, Zhang X, Zhang Q, Huang C, Meng G, Liu W, Wang X.
  ``{EAT-NAS:} elastic architecture transfer for accelerating large-scale neural
  architecture search.'' CoRR abs/1901.05884, 2019.

\bibitem{b4}
Jin H, Song Q, Hu X,``Auto-keras: An efficient neural architecture search
  system.'' In: Teredesai A, Kumar V, Li Y, Rosales R, Terzi E, Karypis G (eds)
  Proceedings of the 25th {ACM} {SIGKDD} International Conference on Knowledge
  Discovery {\&} Data Mining, {KDD} 2019, Anchorage, AK, USA, August 4-8, 2019,
  {ACM}, pp 1946--1956, 2019. DOI:10.1145/3292500.3330648.

\bibitem{b5}
Zoph B, Le QV, ``Neural architecture search with reinforcement learning.'' 2017.

\bibitem{b6}
Zhou ZH, ``Ensemble methods: foundations and algorithms.'' Chapman and
  Hall/CRC. 2012.

\bibitem{b7}
 (2017) 5th International Conference on Learning Representations, {ICLR} 2017,
  Toulon, France, April 24-26, 2017, Conference Track Proceedings,
  OpenReview.net,

\bibitem{b8}
Weill C, Gonzalvo J, Kuznetsov V, Yang S, Yak S, Mazzawi H, Hotaj E, Jerfel G,
  Macko V, Adlam B, Mohri M, Cortes C, ``Adanet: A scalable and flexible
  framework for automatically learning ensembles.'' 1905.00080, 2019.

\bibitem{b9}
Bengio Y, Practical recommendations for gradient-based training of deep
  architectures. In: ``Neural networks: Tricks of the trade'', Springer, pp
  437--478, 2012.

\bibitem{b10}
Webb GI, Zheng Z, ``Multistrategy ensemble learning: Reducing error by
  combining ensemble learning techniques.'' IEEE Transactions on Knowledge and
  Data Engineering 16(8):980--991, 2004.

\bibitem{b11}
Li H, Xu Z, Taylor G, Studer C, Goldstein T, ``Visualizing the loss
  landscape of neural nets.'' In: Advances in Neural Information Processing
  Systems, pp 6389--6399, 2018.

\bibitem{b12}
Goodfellow IJ, Vinyals O, ``Qualitatively characterizing neural network
  optimization problems.'' In: Bengio Y, LeCun Y (eds) 3rd International
  Conference on Learning Representations, {ICLR} 2015, San Diego, CA, USA, May
  7-9, 2015, Conference Track Proceedings, http://arxiv.org/abs/1412.6544.

\bibitem{b13}
Huang G, Li Y, Pleiss G, Liu Z, Hopcroft JE, Weinberger KQ, ``Snapshot
  ensembles: Train 1, get {M} for free.'' In:  \cite{b7}, https://openreview.net/forum?id=BJYwwY9ll, 2017.

\bibitem{b14}
Smith LN, ``Cyclical learning rates for training neural networks.'' Computer
  Science pp 464--472, 2015.

\bibitem{b15}
Xie J, Xu B, Zhang C, ``Horizontal and vertical ensemble with deep
  representation for classification.'' CoRR abs/1306.2759,
http://arxiv.org/abs/1306.2759, 2013.

\bibitem{b16}
Chen H, Lundberg S, Lee S, ``Checkpoint ensembles: Ensemble methods from a
  single training process.'' CoRR abs/1710.03282,
http://arxiv.org/abs/1710.03282, 2017.

\bibitem{b17}
Loshchilov I, Hutter F, ``{SGDR:} stochastic gradient descent with warm
  restarts.'' 2017. 

\bibitem{b18}
Wen L, Gao L, Li X, ``A new snapshot ensemble convolutional neural network
  for fault diagnosis.'' IEEE Access 7:32037--32047, 2019.

\bibitem{b19}
Garipov T, Izmailov P, Podoprikhin D, Vetrov DP, Wilson AG, ``Loss
  surfaces, mode connectivity, and fast ensembling of dnns.'' In: Bengio S,
  Wallach H, Larochelle H, Grauman K, Cesa-Bianchi N, Garnett R (eds) Advances
  in Neural Information Processing Systems 31, Curran Associates, Inc., pp
  8789--8798, http://papers.nips.cc/paper/8095-loss-surfaces-mode-connectivity-and-fast-ensembling-of-dnns.pdf, 2018.

\bibitem{b20}
Inoue H, ``Adaptive ensemble prediction for deep neural networks based on
  confidence level.'' In: Chaudhuri K, Sugiyama M (eds) The 22nd International
  Conference on Artificial Intelligence and Statistics, {AISTATS} 2019, 16-18
  April 2019, Naha, Okinawa, Japan, {PMLR}, Proceedings of Machine Learning
  Research, vol~89, pp 1284--1293,
http://proceedings.mlr.press/v89/inoue19a.html, 2019.

\bibitem{b21}
Ju C, Bibaut A, van~der Laan M, ``The relative performance of ensemble
  methods with deep convolutional neural networks for image classification.''
  Journal of Applied Statistics 45(15):2800--2818, 2018.

\bibitem{b22}
Krizhevsky A, Hinton G, ``Learning multiple layers of features from tiny
  images.'' Computer Science Department, University of Toronto, Tech Rep 1, 2009.

\bibitem{b23}
Lake B, Salakhutdinov R, Gross J, Tenenbaum J, ``One shot learning of
  simple visual concepts.'' In: Proceedings of the annual meeting of the
  cognitive science society, vol~33, 2011.

\bibitem{b24}
He K, Zhang X, Ren S, Jian S, ``Deep residual learning for image
  recognition.'' In: 2016 IEEE Conference on Computer Vision and Pattern
  Recognition (CVPR), 2016.

\bibitem{b25}
Zagoruyko S, Komodakis N, ``Wide residual networks.'' In: Wilson RC, Hancock
  ER, Smith WAP (eds) Proceedings of the British Machine Vision Conference
  2016, {BMVC} 2016, York, UK, September 19-22, 2016, {BMVA} Press,
http://www.bmva.org/bmvc/2016/papers/paper087/index.html, 2016.

\bibitem{b26}
Simonyan K, Zisserman A. ``Very deep convolutional networks for large-scale
  image recognition.'' Computer Science, 2014.

\bibitem{b27}
Koch G, Zemel R, Salakhutdinov R, ``Siamese neural networks for one-shot
  image recognition.'' In: ICML deep learning workshop, vol~2, 2015.

\bibitem{b28}
Vinyals O, Blundell C, Lillicrap T, Wierstra D, et~al, ``Matching networks
  for one shot learning.'' In: Advances in neural information processing systems,
  pp 3630--3638, 2016.
\end{thebibliography}
\end{document}